\documentclass[11pt]{article}

\usepackage[preprint]{acl}
\usepackage{times}
\usepackage{latexsym}

\usepackage[T1]{fontenc}    
\usepackage[utf8]{inputenc} 

\usepackage{microtype}
\usepackage{inconsolata}    

\usepackage{graphicx}
\usepackage{subcaption}

\usepackage{booktabs}

\usepackage{amsfonts}
\usepackage{nicefrac}

\usepackage{xcolor}

\usepackage{hyperref}
\usepackage{url}

\usepackage{listings}

\usepackage{titlesec}

\makeatletter
\renewcommand\paragraph{\@startsection{paragraph}{4}{\z@}%
  {1.0ex plus 0.5ex minus .2ex}
  {0.5ex plus 0.2ex}
  {\normalfont\normalsize\bfseries}} 
\makeatother

\hypersetup{bookmarksdepth=4} 

\definecolor{string}{RGB}{42,0,255}
\definecolor{number}{RGB}{128,0,128}
\definecolor{boolean}{RGB}{0,128,0}
\definecolor{null}{RGB}{128,128,128}
\definecolor{key}{RGB}{163,21,21}

\lstdefinelanguage{json}{
    basicstyle=\ttfamily\small,
    numbers=left,
    numberstyle=\tiny,
    stepnumber=1,
    numbersep=8pt,
    showstringspaces=false,
    breaklines=true,
    literate=
     *{true}{{{\color{boolean}true}}}{4}
      {false}{{{\color{boolean}false}}}{5}
      {null}{{{\color{null}null}}}{4}
      {"}{{{\color{string}"}}}{1}
      {:}{{{\color{black}:}}}{1}
      {,}{{{\color{black},}}}{1}
      {0}{{{\color{number}0}}}{1}
      {1}{{{\color{number}1}}}{1}
      {2}{{{\color{number}2}}}{1}
      {3}{{{\color{number}3}}}{1}
      {4}{{{\color{number}4}}}{1}
      {5}{{{\color{number}5}}}{1}
      {6}{{{\color{number}6}}}{1}
      {7}{{{\color{number}7}}}{1}
      {8}{{{\color{number}8}}}{1}
      {9}{{{\color{number}9}}}{1}
}

\title{OleSpeech-IV: A Large-Scale Multispeaker and Multilingual Conversational Speech Dataset with Diverse Topics}

\author{%
  \begin{minipage}{\textwidth}\centering
\small
\textbf{Wei Chu}$^{1}$, \textbf{Yuanzhe Dong}$^{1,2}$, 
\textbf{Ke Tan}$^{1}$, \textbf{Dong Han}$^{1}$, \textbf{Xavier Menendez-Pidal}$^{1}$ \\
{\normalfont $^{1}$Olewave, San Francisco, USA \quad
$^{2}$Stanford University}
\end{minipage}
  \\[2ex] 
  \begin{minipage}[t]{0.2\textwidth}
    \centering
    \small
    \textbf{Ruchao Fan}$^{4}$ \\
    {\normalfont $^{4}$Microsoft}
  \end{minipage}%
  \hspace{0.01\textwidth}
  \begin{minipage}[t]{0.2\textwidth}
    \centering
    \small
    \textbf{Chenfeng Miao}$^{5}$ \\
    {\normalfont $^{5}$PingAn Technology}
  \end{minipage}%
  \hspace{0.01\textwidth}
  \begin{minipage}[t]{0.2\textwidth}
    \centering
    \small
    \textbf{Chanwoo Kim}$^{6}$ \\
    {\normalfont $^{6}$Korea University}
   \end{minipage}
   \hspace{0.01\textwidth}
  \begin{minipage}[t]{0.36\textwidth}
  \centering
    \small
    \textbf{Bhiksha Raj}$^{1,7}$, \textbf{Rita Singh}$^{1,7}$ \\
    {\normalfont $^{7}$Carnegie Mellon University}
  \end{minipage}%
  \thanks{Contact: info@olewave.com, please cite this work as: Olewave, “OleSpeech-IV: A Large-Scale Multispeaker and Multilingual Conversational
Speech Dataset with Diverse Topics.” arXiv preprint, arXiv:arXiv\_ID\_TBD, 2025.}
}

\begin{document}

\maketitle
\begin{abstract}
OleSpeech-IV dataset is a large-scale multispeaker and multilingual conversational speech dataset with diverse topics. The audio content comes from publicly-available English podcasts, talk shows, teleconferences, and other conversations. Speaker names, turns, and transcripts are human-sourced and refined by a proprietary pipeline, while additional information such as timestamps and confidence scores is derived from the pipeline. The IV denotes its position as Tier IV in the Olewave dataset series. In addition, we have open-sourced a subset, OleSpeech-IV-2025-EN-AR-100, for non-commercial research use.
\end{abstract}

\section{Introduction}

The creation of a large-scale dataset typically involves two critical stages: \textit{curation} and \textit{cleaning}. Each stage presents unique challenges that must be addressed to ensure both the quality and utility of the resulting dataset.

\subsection{Challenges in curating large-scale speech datasets}
Curating large-scale speech datasets is inherently complex. In this work, we aim to expand the discussion by providing a historical overview of existing large-scale speech resources, while also highlighting the motivation and initiative behind the development of OleSpeech-IV.

\subsection{Challenges in cleaning large-scale speech datasets}
Speech data cleaning is particularly challenging due to the multifaceted nature of speech signals, which introduce complexities beyond those encountered in traditional text-based Natural Language Processing (NLP). While NLP deals with variations in language, vocabulary, and topic, speech processing must additionally contend with channel variations and speaker variations, each of which introduces unique difficulties.

Channel variations refer to differences in the acoustic environment and recording conditions, which can significantly degrade the quality of speech data. These variations include:
\begin{itemize}
    \item Background noise: Environmental sounds, such as traffic, wind, or crowd noise, can obscure speech signals.
    \item Reverberation: Echoes caused by recording in large or reflective spaces can distort the clarity of speech.
    \item Microphone quality: Variations in recording devices can lead to differences in audio fidelity, frequency response, and signal-to-noise ratio (SNR).
    \item Transmission artifacts: In telephony or streaming applications, compression, packet loss, or bandwidth limitations can introduce distortions.
\end{itemize}

Speaker variations add another layer of complexity, as speech signals are highly dependent on individual characteristics, such as:
\begin{itemize}
    \item Accent and dialect: Regional or cultural differences in pronunciation and vocabulary can affect recognition accuracy.
    \item Speaking style: Variations in speaking rate, pitch, volume, and articulation can make it difficult to generalize across speakers.
    \item Emotional state: Stress, excitement, or fatigue can alter speech patterns and introduce variability.
    \item Physiological factors: Age, gender, and vocal tract differences can influence the acoustic properties of speech.
\end{itemize}

\section{OleSpeech-IV Datasets}
OleSpeech-IV dataset is the currently most comprehensive dataset offered by us. The 'IV' means 'Tier IV', and it is necessary for us to explain the Tier I, II, and III datasets, since Tier IV dataset is obtained through filtering these lower tier datasets, and applying more advanced data pipelines.

Available in multiple languages and covering diverse topics, our datasets are organized into four tiers of pre-labeled offerings, each designed to meet varying levels of annotation depth and application requirements.

\subsection{Tier I -- Untranscribed Speech Datasets}
These datasets do not include spoken content transcripts but come with the following labels:

\subsubsection*{Mandatory Metadata}
\begin{itemize}
    \item Language labels (see Appendix: Languages for details)
    \item Topic tags (see Appendix: Topics for details)
\end{itemize}

\subsubsection*{Optional Signal Condition Tags}
\begin{itemize}
    \item Signal-to-Noise Ratio (SNR)
    \item Speech Percentage Ratio
    \item \dots
\end{itemize}

\subsubsection*{Optional Text-Based Tags}
\begin{itemize}
    \item Summaries and keywords
    \item Audio creation metadata
    \item \dots
\end{itemize}

Take English language for example, we currently offer over 1 million hours of audio at this tier and are ready to collect more on demand.

\subsection{Tier II -- Pseudo-Transcribed Speech Datasets}
These datasets include machine-generated transcripts with validation scores:

\subsubsection*{Included Labels}
\begin{itemize}
    \item All labels from Tier I
\end{itemize}

\subsubsection*{Transcript Validation}
\begin{itemize}
    \item Accurate onset and offset timings for each utterance
    \item Confidence scores at both the word and utterance levels
\end{itemize}

Take English language for example, we currently offer over 200,000 hours of audio at this tier and can scale further if needed.

\subsection{Tier III -- Human-in-the-Loop (HITL) Transcribed Speech Datasets}
These datasets contain transcripts created and validated by humans:

\subsubsection*{Included Labels}
\begin{itemize}
    \item All labels from Tier I
\end{itemize}

\subsubsection*{Human-Generated Transcripts}
\begin{itemize}
    \item Validation equivalent to Tier II standards
\end{itemize}

Take English language for example, we currently offer over 40,000 hours of audio at this tier, with the ability to expand upon request.

\subsection{Tier IV -- Advanced-Labeled Speech Datasets}
These are the most comprehensive datasets with enriched annotations:

\subsubsection*{Included Labels}
\begin{itemize}
    \item All data and labels from Tier III
\end{itemize}

\subsubsection*{Additional Advanced Labels}
\begin{itemize}
    \item Speaker labels and timing information for conversational speech (see Appendix: Conversational Speech Datasets)
    \item Translated transcripts of spoken content (optional)
    \item Optional verbatim (word-for-word) transcripts (optional)
    \item ...
    \item Custom labels as specified by clients
\end{itemize}

We currently have over 5,000 hours of audio, and we name it as Olewave-IV-2025, which 2025 denotes the year.

\subsubsection{OleSpeech-IV-2025-EN-AR-100 -- for Non-Commercial Research Use}
The OleSpeech-IV-2025-EN-AR-100 dataset is a subset of the Olewave-IV-2025 collection. It contains 100 hours of English (EN) speech with accents from all regions (AR). The data is divided into training, development, and test sets with an 8:1:1 ratio.

\section{Methodologies}
Olewave’s state-of-the-art and highly customizable speech data cleaning pipeline seamlessly integrates with its speech data collection pipeline, together forming the Olewave’s speech data curation pipeline. This end-to-end system is meticulously designed to deliver high-quality, consistent, and cost-effective speech datasets at scale, tailored for a wide range of downstream applications.

Additionally, we are offering large-scale, cost-effective speech datasets curated through our pipeline, available in multiple languages and covering diverse topics. Figure 1 illustrates the streamlined workflow of our speech data cleaning pipeline, showcasing its ability to transform raw, unstructured speech data into polished, actionable datasets. At the core of this pipeline is Olewave’s signature speech-to-text alignment module, \textit{Olign}, which enables accurate and scalable cleaning of speech data.

\begin{figure*}[]
  \centering
  \includegraphics[width=0.9\linewidth]{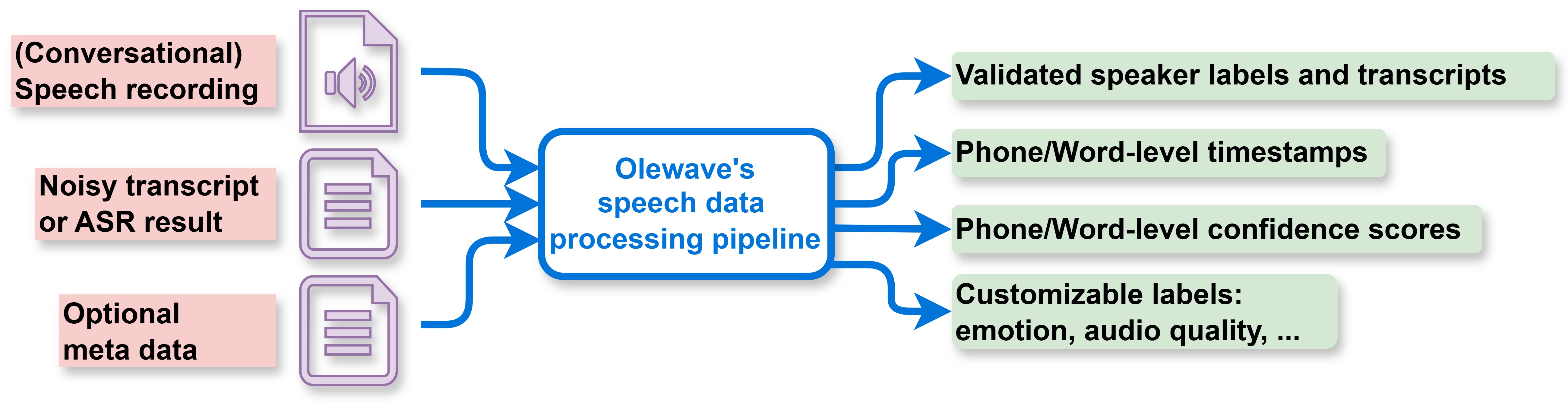}
  \caption{Olewave’s speech data processing pipeline. The central component is the speech-to-text alignment module \textbf{Olign}, which can be accessed through the Olign SDK or deployed on-premises.}
  \label{fig:olewave_data_pipeline}
\end{figure*}

This innovative pipeline takes raw speech recordings, corresponding transcripts, and optional metadata as inputs, and robustly cleans noisy transcriptions and labels. It delivers validated speaker labels and transcripts, and rich features such as reliable word-level timestamps and confidence scores. By leveraging advanced alignment techniques, our pipeline ensures the highest levels of accuracy and usability, making it an indispensable tool for building high-performance but cost-efficient speech-based systems.

Our pipeline offers several advantages:
\begin{itemize}
    \item Effective: It produces validated speaker labels and transcriptions, accurate word timestamps and not-overconfident confidence scores compared to tools like Whisper and other open-source solutions.
    \item Robustness: It handles improvised conversational speech, including scenarios with speaker overlap and transcripts containing ASR errors.
    \item Extensible: It leverages optional metadata and can be upgraded to incorporate additional modality for increased label reliability. The pipeline also supports plug-and-play integration of various Olewave’s or client’s own label tagging models (e.g., emotion, semantics, …).
    \item Efficiency: It runs quickly and is cost-effective, as it requires no or little GPU resources.
\end{itemize}

Additionally, if your goal is to have human annotators refine Automatic Speech Recognition (ASR) results, the output of our pipeline can significantly reduce labeling costs by:
\begin{itemize}
    \item Precisely highlighting words with accurate timing information, enabling annotators to quickly locate and review mumbled or unclear segments without repeatedly listening to the entire audio.
    \item Guiding annotators to focus on words with mid-range confidence scores, minimizing the need to review every word and streamlining the correction process.
\end{itemize}
By providing these targeted insights, our pipeline enhances efficiency, reduces manual effort, and ensures a more cost-effective annotation workflow.

\subsection{Olign: fine-grained, accurate, and sophisticated speech-to-text alignment}

Olign is our robust and proprietary speech-to-text aligner, accessible via the Olign SDK or through on-premises deployment. It currently supports English, with support for Spanish, Japanese, and Mandarin under development.

\subsubsection{Timestamp}

Timestamps indicate the onset and offset times of a given unit within an audio recording. The unit can be a paragraph, utterance, sentence, word, syllable, or phoneme. Obtaining accurate timestamps is particularly challenging when the audio contains noise, accent variations, speaker overlap, or when the transcripts include errors, such as those produced by automatic speech recognition (ASR) systems.

\begin{figure*}[t] 
  \centering

  \begin{subfigure}{0.9\linewidth}
    \centering
    \includegraphics[width=\linewidth]{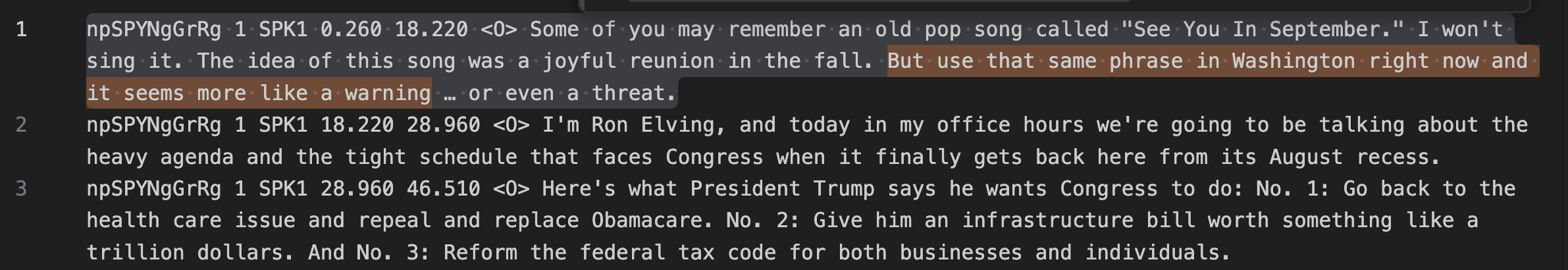}
    \caption{End-to-End-based alignment result}
    \label{fig:e2e_based_align_result}
  \end{subfigure}

  \vspace{1em} 

  \begin{subfigure}{0.9\linewidth}
    \centering
    \includegraphics[width=\linewidth]{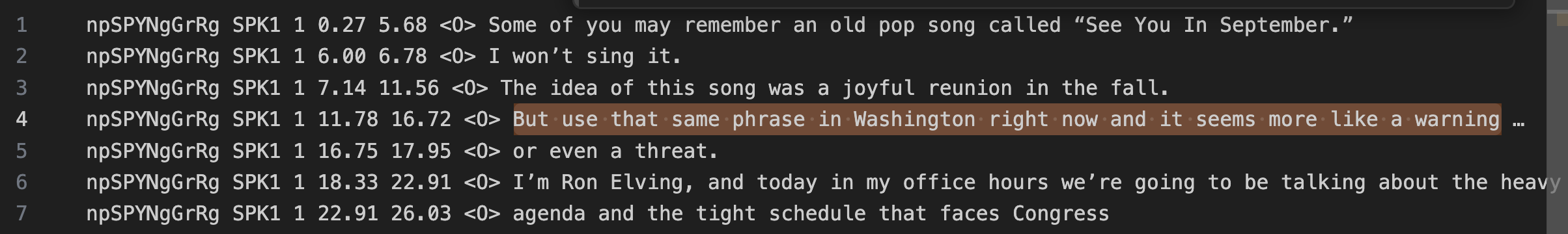}
    \caption{Olign alignment result (not End-to-End)}
    \label{fig:olign_align}
  \end{subfigure}

  \caption{Comparison between End-to-End-based alignment (a) and Olign alignment (b) on an audio of 3 minutes. Only the . We haven’t benchmark our accuracy yet (work in progress), but we’ve examined thousands of timestamps and fixed issues.}
  \label{fig:stm_compare_e2e_vs_olign}
\end{figure*}

\paragraph{Timestamp Granularity}
As shown in Figure~\ref{fig:stm_compare_e2e_vs_olign}, we compared alignment results from End-to-End (E2E) aligner and Olign when having an audio recording with its corresponding transcript as input. Olign produces finer, sentence-level timestamps, whereas E2E alignment outputs coarser, chunk-level timestamps. Word-level timestamps generated by Olign are not shown here. It is worth noting that the E2E alignment was configured to process the audio in fixed chunks of no more than 30 seconds to accommodate GPU memory constraints. In our view, it is preferable to segment the audio at sentence-ending punctuation, as this preserves semantic coherence within each segment.

Olign is capable of producing timestamps at the phone level, and we plan to provide further details on this capability in the future.

\paragraph{Timestamp Accuracy}
While the End-to-End (E2E) aligner can be configured for finer-grained alignment, we have observed that it often produces boundary errors, making this configuration inadvisable. 

\begin{figure*}[t] 
  \centering

  \begin{subfigure}{0.9\linewidth}
    \centering
    \includegraphics[width=\linewidth]{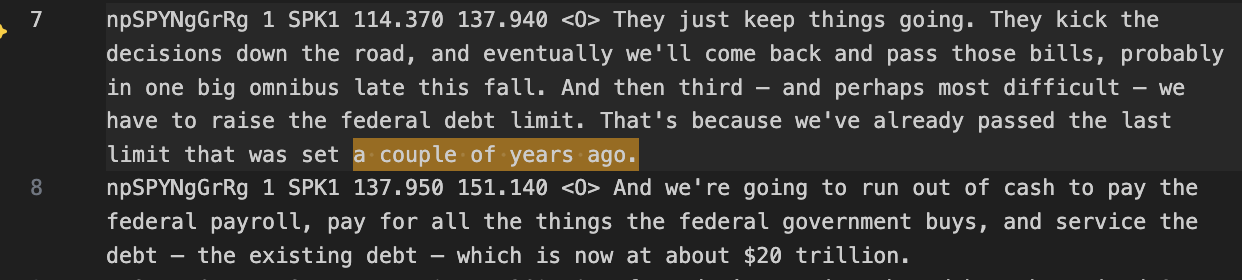}
    \caption{End-to-End alignment result.}
    \label{fig:e2e_align_result}
  \end{subfigure}

  \vspace{1em} 

  \begin{subfigure}{0.9\linewidth}
    \centering
    \includegraphics[width=\linewidth]{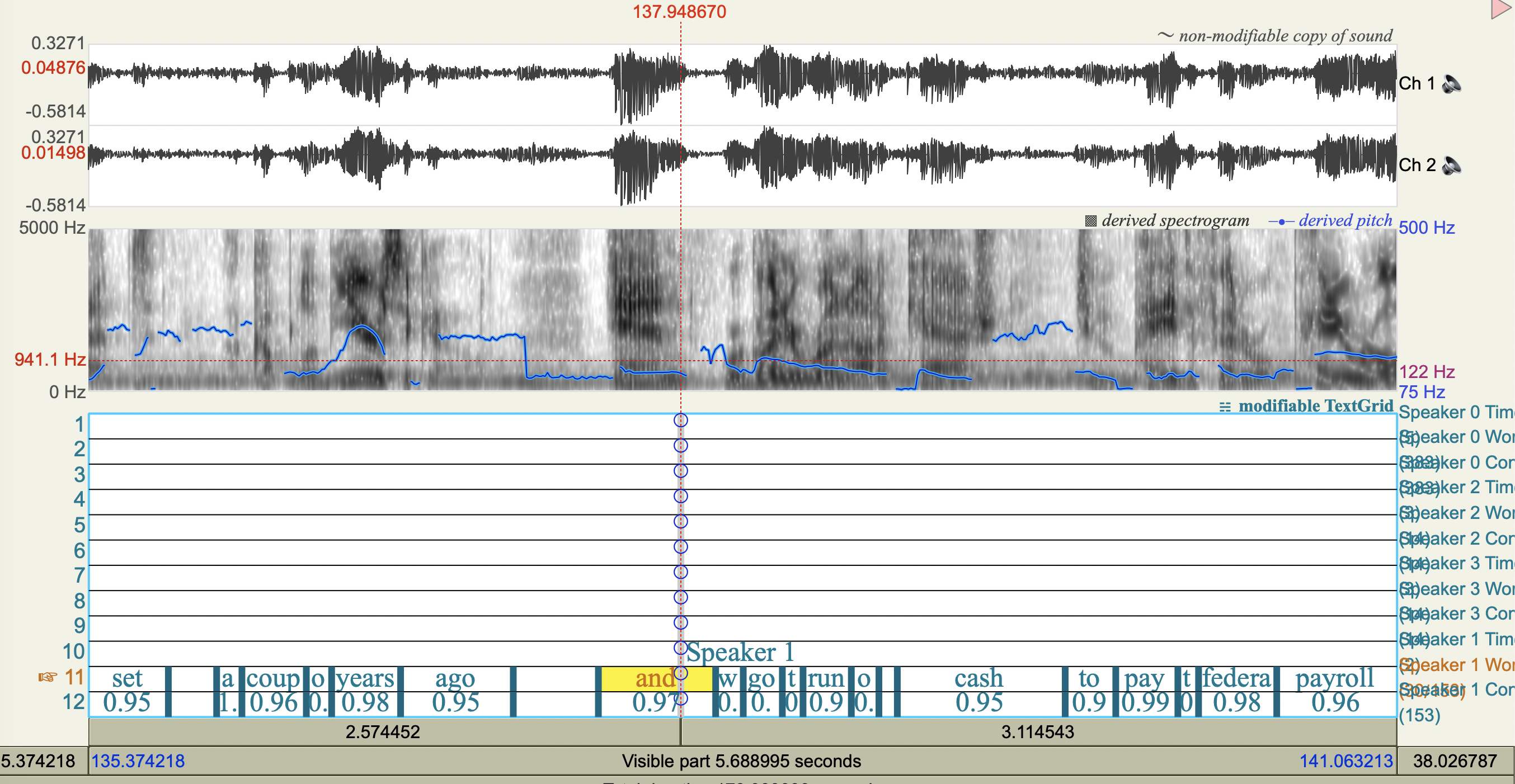}
    \caption{In the E2E result, the word “ago” is incorrectly aligned: its ending time is set to 137.94s, overlapping with the start of the next word “end” (red vertical line). Olign provides the correct ending time for “ago.”. The number below each word is its confidence score.}
    \label{fig:olign_correct_align}
  \end{subfigure}

  \caption{Explaining alignment accuracy issue of an E2E aligner.}
  \label{fig:e2e_vs_olign}
\end{figure*}

As illustrated in Figure~\ref{fig:e2e_vs_olign}, we observed that the E2E model likes to place the end of a segment exactly at the start of the first word of the following segment. In cases with long silences between segments, this can lead either to the inclusion of excessive silence at the end of the previous segment or to misaligned boundaries.

Olign is designed to handle ASR outputs containing errors. For instance, Whisper often generates numerous insertion errors, sometimes referred to as “hallucinations.” In such cases, Olign typically assigns a zero duration to these inserted words while still providing precise timestamps for correctly recognized words. We plan to elaborate on this functionality in future work.

\begin{figure*}[t] 
  \centering
  \includegraphics[width=0.9\linewidth]{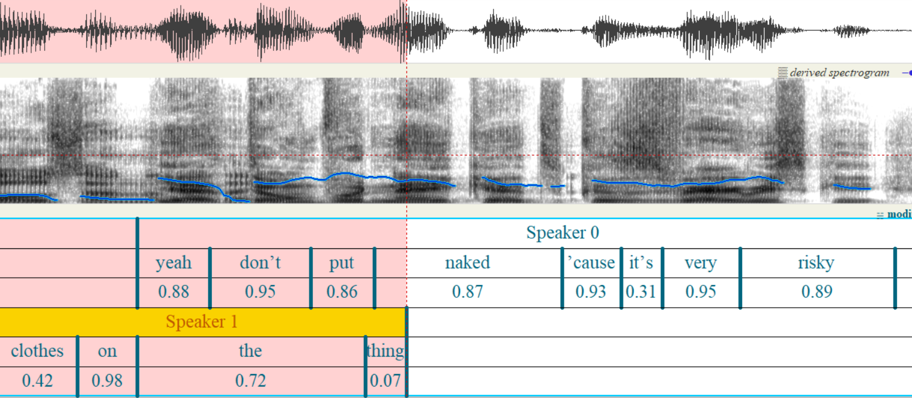}
  \caption{Example demonstrating Olign aligning overlapping speech from two speakers. The number below each word indicates its confidence score.}
  \label{fig:olign_overlap}
\end{figure*}

\paragraph{Speaker Overlapping}
Speaker overlap is common in conversational speech, particularly during turn-taking or backchannel responses, and poses a challenge for speech-to-text alignment. Olign can handle transcripts with speaker tags and generate word-level timestamps even in overlapping segments. Figure~\ref{fig:olign_overlap} illustrates an example. (Benchmarking and further improvements are ongoing.)

In summary, Olign provides:  

\begin{itemize}
    \item \textbf{Sentence- and word-level timestamps:} Tight sentence boundaries are particularly useful for training conversational speech models with low response latency. Note that phone-level timestamps are available upon request.
    \item \textbf{Confidence scores at both word and sentence levels:} Low-confidence segments can be excluded from training, reducing noise and improving overall model quality.  
    \item \textbf{Alignment with speaker information:} Facilitates training for speaker diarization and conversational speech models.
\end{itemize}

\begin{figure*}[t] 
  \centering
  \includegraphics[width=0.9\linewidth]{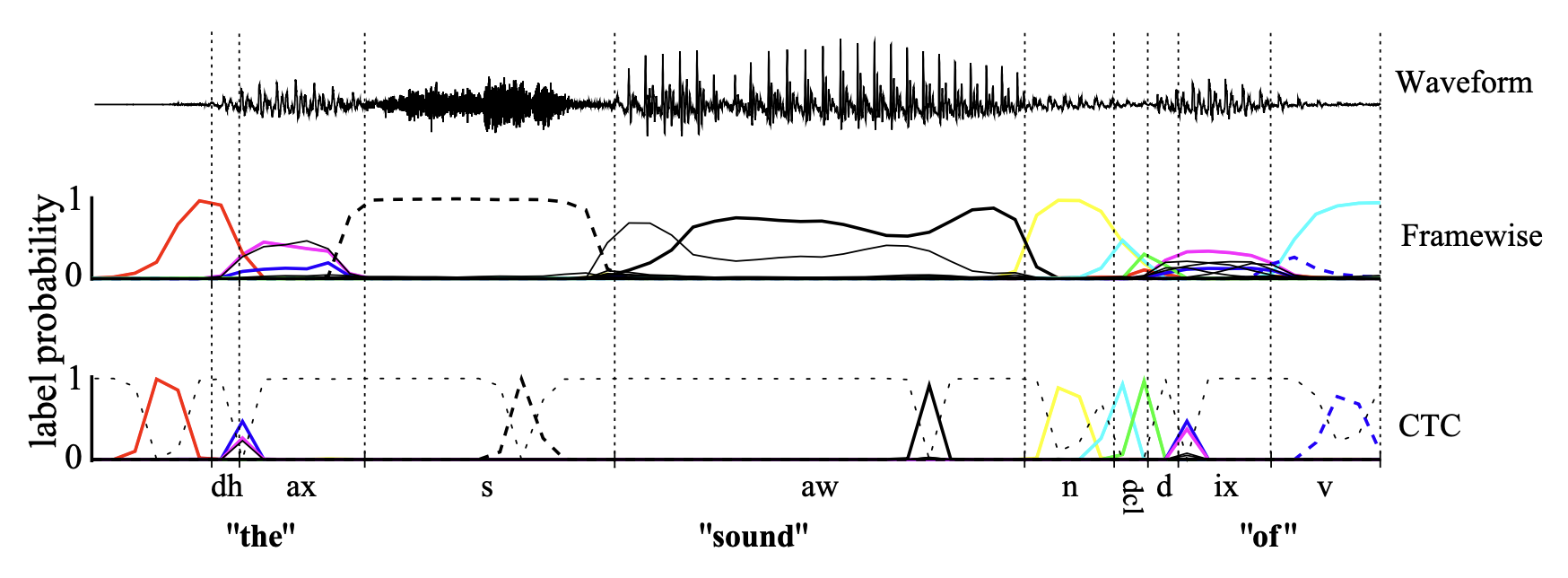}
  \caption{Caption adapted from the CTC paper\cite{graves2006ctc}: "Framewise and CTC networks classifying a speech signal. The shaded lines are the output activations, corresponding to the probabilities of observing phonemes at particular times. The CTC network predicts only the sequence of phonemes (typically as a series of spikes, separated by ‘blanks’, or null predictions), while the framewise network attempts to align them with the manual segmentation (vertical lines). The framewise network receives an error
for misaligning the segment boundaries, even if it predicts the correct phoneme (e.g. ‘dh’). When one phoneme always occurs beside another (e.g. the closure ‘dcl’ with the stop ‘d’), CTC tends to predict them together in a double spike.
The choice of labelling can be read directly from the CTC outputs (follow the spikes), whereas the predictions of the framewise network must be post-processed before use." Figure is also adapted from \cite{graves2006ctc}.}
  \label{fig:ag_ctc_align}
\end{figure*}

\begin{figure*}[t] 
  \centering
  \centering
  \includegraphics[width=0.9\linewidth]{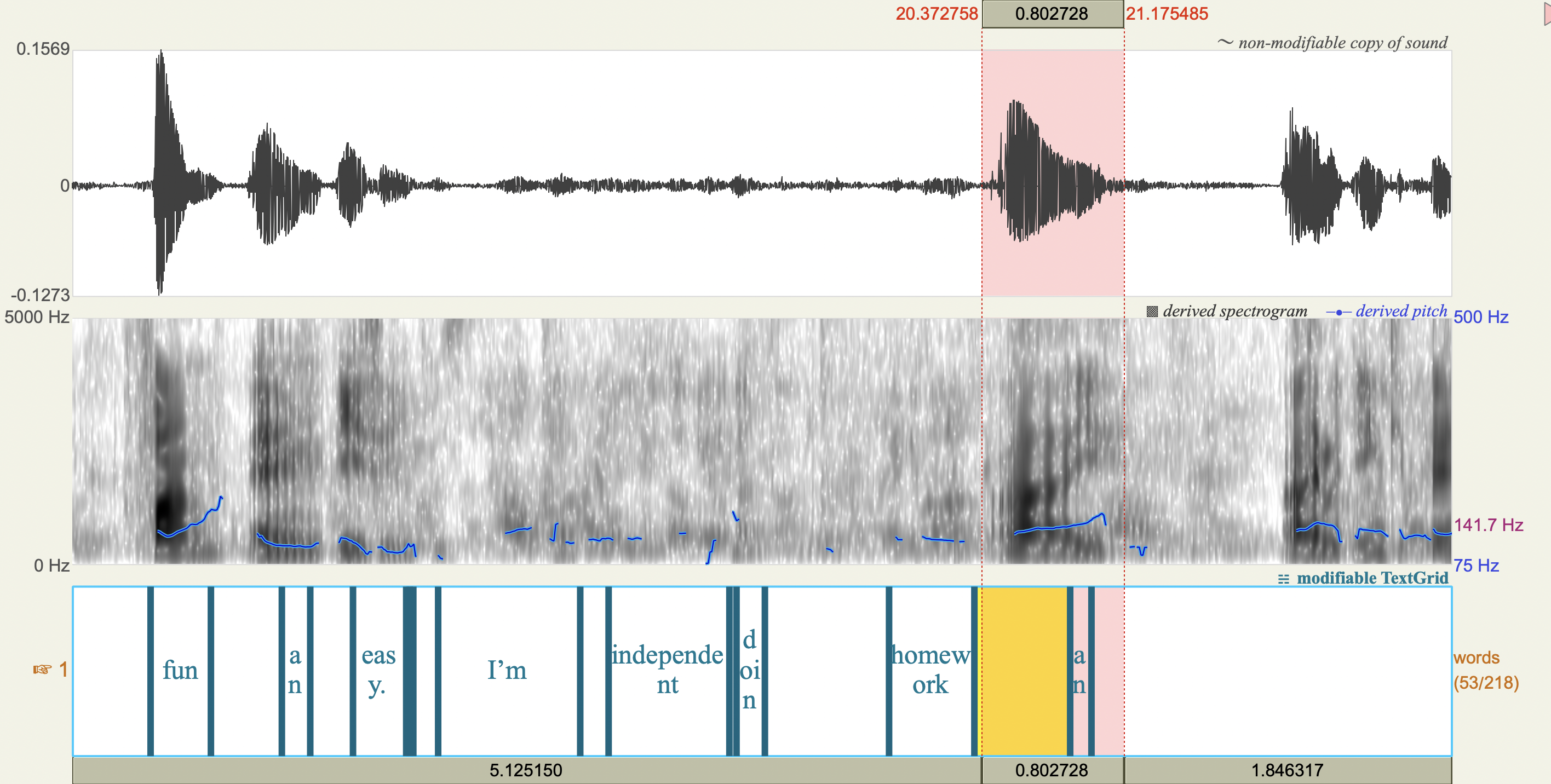}
  \caption{Caption adapted from WhisperX's issue \#297~\cite{bain2023whisperx}: "... Sometimes the transcriptions are all shifted significantly forward in time. So that the transcription occurs seconds before the speech. This usually adjusts itself later in the transcript. In this screenshot, the word "I'm" appears in the transcript at the position displayed at the bottom of the screen, but in the audio it is not spoken until the highlighted section on the right. Before that, there is only low background noise. This seems to happen mostly in moments of no speech. I wonder if the way the audio is cut up before processing is causing this." Figure is also adapted from \cite{bain2023whisperx}}
  \label{fig:whisperx_error_align}
\end{figure*}

\paragraph{Further read}
\label{sec:avoid_e2e_align}
When employing End-to-End (E2E) frameworks for speech-to-text alignment using mechanisms like attention weights or CTC alignment, it is common to encounter misalignments. This is because these models are primarily optimized to predict the most probable sequence of wordpieces or tokens, rather than ensuring precise temporal alignment between the audio and the corresponding text. The attention weights or CTC activations, such as the spike triggers, do not inherently guarantee accurate timestamp alignment with the actual speech~\cite{graves2006ctc}~\cite{bain2023whisperx}. Figure~\ref{fig:ag_ctc_align} Graves’ seminal CTC paper illustrates that CTC-based networks are inherently unable to produce precise phone- or word-level timestamps. Figure~\ref{fig:whisperx_error_align} from WhisperX's issue \#297~\cite{bain2023whisperx} shows that attention weights-based methods are also prone to significant alignment errors.  In contrast, Olign provides more advanced alignment than these aligners, as they avoid smearing segment boundaries.

\subsubsection{Confidence Score}

Confidence scores at the word or sentence level are valuable for validating transcripts, particularly when the data originates from public sources or from error-prone ASR systems.

However, large end-to-end ASR models, such as Whisper, often exhibit overconfidence: they assign high posterior probabilities (e.g., above 0.8) to recognized words even when the recognition is incorrect. This behavior occurs because these models are optimized for sequence-level accuracy rather than calibrated per-token confidence estimates. Consequently, the resulting confidence scores may not reliably reflect the true likelihood of correctness, especially in cases of misrecognitions or ambiguous speech segments.

We show some low confidence score cases in the following Figure~\ref{fig:olign_confidence_score}.

\begin{figure*}[t] 
  \centering
  \includegraphics[width=0.9\linewidth]{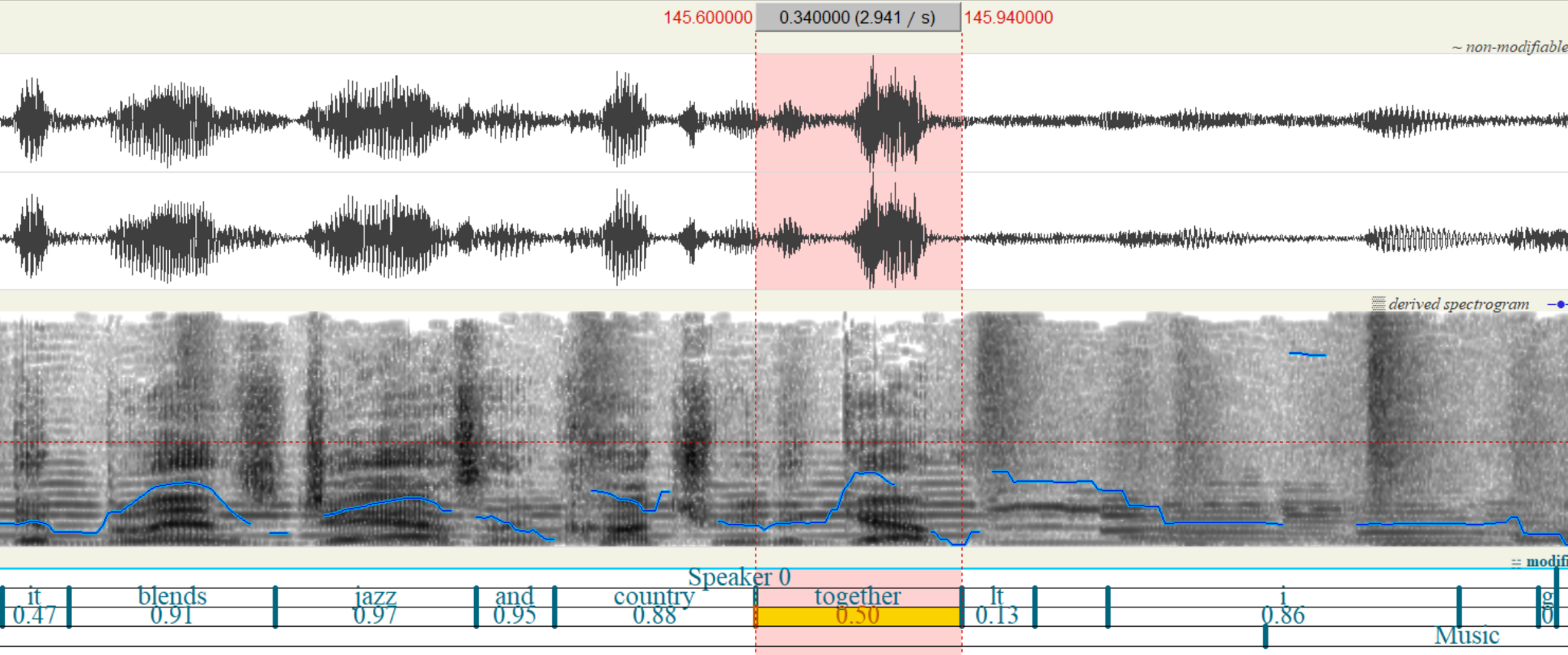}
  \caption{An example of the postprocessing of noisy spontaneous speech is shown, where the number beneath each word represents its confidence score. Some explanations of these scores are as follows: \textbf{a)} The word "together": It was pronounced like "togeth-", with the ending "-er" missing. The "-o-" vowel is very short in duration and almost imperceptible.\\
  \textbf{b)} The starting word "it": This word is barely audible, although a fricative phone is visible on the spectrogram. Its low confidence score of 0.47 indicates that the word is likely missing.\\
  \textbf{c)} The inserted word ``lt'' at the end: This word was not actually spoken; the transcript erroneously includes ``\texttt{\&lt;}'' (the ``<'' symbol used in HTML). Its very low score of 0.13 indicates an inserted word that can be removed by our data cleaning pipeline.
}
  \label{fig:olign_confidence_score}
\end{figure*}

\section{Conclusion}

In this work, we present \textbf{OleSpeech-IV}, a large-scale multilingual and multispeaker conversational speech dataset featuring diverse topics and advanced annotations. Developed through a tiered curation framework spanning OleSpeech-I, II, III, and IV, the collection provides researchers and practitioners with resources at varying levels of transcription and labeling, from raw untranscribed speech to richly annotated conversations with detailed speaker and timing information.  

We also described \textbf{Olign}, our proprietary speech-to-text alignment module, which enables fine-grained timestamps, robust handling of overlapping speech, and calibrated confidence scoring—capabilities that address common shortcomings in end-to-end alignment systems. Together with our speech data curation pipeline, Olign ensures the production of high-quality, scalable, and cost-effective resources suitable for a broad spectrum of downstream speech technologies.  

By releasing a subset of OleSpeech-IV—\textbf{OleSpeech-IV-2025-EN-AR-100}—for non-commercial research, we aim to lower the barrier to academic exploration and benchmarking in multilingual conversational speech processing. We believe that OleSpeech-IV and the methodologies behind it will help advance speech recognition, diarization, alignment, and broader spoken language understanding, ultimately enabling the community to develop more reliable and inclusive speech systems.

\section{Acknowledgments}

We sincerely thank our clients for their valuable feedback and thoughtful suggestions, though we are unable to mention them by name due to mutual NDA agreements. We also greatly appreciate the constructive discussions with researchers, as well as the engaging conversations we shared with those who expressed interest in our work during our exhibitions at Interspeech and ICASSP.

\bibliography{main} 











\section*{Appendix}   

\setcounter{section}{4} 
\setcounter{subsection}{0}  

\subsection{Supported Languages}
The dataset includes languages spoken in most G20 countries and Russia:

\begin{itemize}
    \item \textbf{English} (United Kingdom, USA, Canada, Australia, India, South Africa)
    \item \textbf{Chinese} (China)
    \item \textbf{Japanese} (Japan)
    \item \textbf{French} (France, Canada)
    \item \textbf{Spanish} (Spain, Mexico, Argentina, Latin America)
    \item \textbf{German} (Germany)
    \item \textbf{Italian} (Italy)
    \item \textbf{Russian} (Russia)
    \item \textbf{Korean} (South Korea)
    \item \textbf{Arabic} (Saudi Arabia, Egypt)
    \item \textbf{Portuguese} (Portugal, Brazil)
    \item \textbf{Hindi} (India)
\end{itemize}

We can also provide dialects of certain spoken languages, e.g., UK English or European vs.\ Latin American Spanish.

\subsection{Supported Topics}
The topics include:

\begin{itemize}
    \item Film \& Animation
    \item Autos \& Vehicles
    \item Music
    \item Pets \& Animals
    \item Sports
    \item Travel \& Events
    \item Gaming
    \item People \& Blogs
    \item Comedy
    \item Entertainment
    \item News \& Politics
    \item How-to \& Style
    \item Education
    \item Science \& Technology
    \item Nonprofits \& Activism
\end{itemize}

We can assign advanced topic tags based on specifications provided by the client.

\subsection{Data Format of Multi-Speaker Conversational Speech Datasets}
\begin{figure*}[t]
  \centering
  \includegraphics[width=0.9\linewidth]{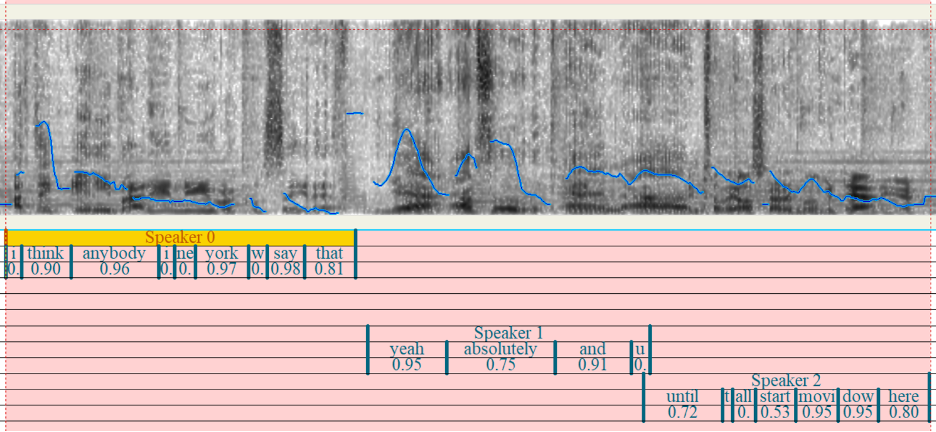}
  \caption{The conversation is from real-world interactions, not synthetic or prompted speech. The transcripts were manually uploaded by humans, not generated by ASR. Speaker labels were also manually added by humans, not derived from speaker diariazation algorithms. The accompanying JSON file includes detailed transcriptions, with speaker time intervals, conversation transcripts, word-level timestamps, and confidence scores. Phone-level timestamps and confidence scores are available upon request too.}
  \label{fig:conversation-snippet}
\end{figure*}

Figure~\ref{fig:conversation-snippet} provides a visual example with labels for three speakers, including word-level timestamps and confidence scores. The sample can be accessed on Google Drive at the following location: \href{https://drive.google.com/drive/folders/1PGdR4KHTyg9rg1TVmb4NhqYhEcHnFsA-}\texttt{https://drive.google.com/drive/folders/\\
1PGdR4KHTyg9rg1TVmb4NhqYhEcHnFsA-}

The audio recordings are raw and unprocessed. In most cases, they contain two channels with the same content. We do not apply denoising, channel separation, or segmentation.

Each audio recording is paired with a set of labels provided in JSON format. The JSON object includes two top-level keys: \verb|version| and \verb|data|. The \verb|data| field follows the structure shown in the example.

\begin{itemize}
    \item \verb|start_time| and \verb|end_time| are expressed in seconds.
    \item The \verb|block| field contains only one speaker, and may contain one or more segments.
    \item Each segment has:
    \begin{itemize}
        \item its own start and end times,
        \item a transcript (we are still improving the text normalization),
        \item a confidence score at the segment level,
        \item and an indicator of whether it overlaps with an adjacent segment from another speaker.
    \end{itemize}
\end{itemize}
For instance, speaker 0 speaks from 264.38s to 264.32s while speaker 1 speaks from 264.38s to 270.78s, resulting in an overlap.

\begin{lstlisting}[language=json,caption={Conversation snippet from \texttt{...en.spkr.block.json}}]
{
  "speaker": 0,
  "start_time": "0.05",
  "end_time": "37.80",
  "confidence": "0.88",
  "block": [
    {
      "start_time": "0.05",
      "end_time": "2.08",
      "transcript": " that he got there,",
      "confidence": "0.84",
      "overlap": false
    },
    {
      "start_time": "2.14",
      "end_time": "4.43",
      "transcript": "that he got to the place to act like a fighter,",
      "confidence": "0.93",
      "overlap": false
    },
    // ...
    {
      "start_time": "34.23",
      "end_time": "37.80",
      "transcript": "this is Cus, with punches with bad intentions.",
      "confidence": "0.81",
      "overlap": false
    }
  ]
},
// ...
{
        "speaker": 1,
        "start_time": "264.38",
        "end_time": "264.62",
        "confidence": "0.80",
        "block": [
            {
                "start_time": "264.38",
                "end_time": "264.62",
                "transcript": "Yeah.",
                "confidence": "0.80",
                "overlap": true
            }
        ]
    },
    {
        "speaker": 0,
        "start_time": "264.38",
        "end_time": "270.78",
        "confidence": "0.74",
        "block": [
            {
                "start_time": "264.38",
                "end_time": "265.10",
                "transcript": "You know what I mean?",
                "confidence": "0.54",
                "overlap": true
            },
            // ...
    }
// ...
}
\end{lstlisting}

In addition, word-level information is included, as illustrated in the following example. The \texttt{word\_details} field provides a list of entries, each containing the word, its start and end times, and a confidence score.

\begin{lstlisting}[language=json,caption={Conversation snippet from \texttt{...en.spkr.wd.block.json}}]
{
    "speaker": 0,
    "start_time": "0.05",
    "end_time": "37.80",
    "confidence": "0.88",
    "block": [
        {
            "start_time": "0.05",
            "end_time": "2.08",
            "transcript": " that he got there,",
            "overlap": false,
            "word_details": [
                {
                    "start_time": "0.05",
                    "end_time": "1.00",
                    "word": "that",
                    "confidence": "0.56"
                },
                {
                    "start_time": "1.36",
                    "end_time": "1.56",
                    "word": "he",
                    "confidence": "0.89"
                },
                {
                    "start_time": "1.56",
                    "end_time": "1.90",
                    "word": "got",
                    "confidence": "0.98"
                },
                {
                    "start_time": "1.90",
                    "end_time": "2.08",
                    "word": "there",
                    "confidence": "0.93"
                }
            ]
        },
    // ...
    ]
// ...
}
\end{lstlisting}

\end{document}